\documentclass[journal]{IEEEtran}
\usepackage[pdftex]{graphicx}
\usepackage{epsfig}
\usepackage[cmex10]{amsmath}
\usepackage{tabularx, makecell, multirow, array, multicol}
\usepackage{rotating}

\usepackage[dvipsnames]{xcolor}
\usepackage{pgfplots}
\usepackage{tikz}
\usetikzlibrary{positioning}
\usepackage{subfigure}
\pgfplotsset{compat=1.14}
\usepackage{url}

\title{Landmine Detection Using Autoencoders on Multi-polarization GPR Volumetric Data}

\author{Paolo~Bestagini,~\IEEEmembership{Member,~IEEE,}
        Federico~Lombardi,~\IEEEmembership{Student Member,~IEEE,}        
        Maurizio~Lualdi,
       	Francesco~Picetti,~\IEEEmembership{Student Member,~IEEE,}
        and~Stefano~Tubaro,~\IEEEmembership{Senior~Member,~IEEE}
    \thanks{P. Bestagini, F. Picetti and S. Tubaro are with the Dipartimento di Elettronica, Informazione e Bioingegneria, Politecnico di Milano - Milano, Italy (email: francesco.picetti / paolo.bestagini / stefano.tubaro@polimi.it)}
    \thanks{F. Lombardi and M. Lualdi are with the Dipartimento di Ingegneria Civile e Ambientale, Politecnico di Milano - Milano, Italy (email: federico.lombardi / maurizio.lualdi@polimi.it)}
	\thanks{This work has been partially supported by the project PoliMIne (Humanitarian Demining GPR System), funded by Polisocial Award from Politecnico di Milano, Milan, Italy.}
	}

\def\E{\mathcal{E}}
\def\D{\mathcal{D}}

\def\v{\mathbf{v}}
\def\vh{\hat{\mathbf{v}}}
\def\h{\mathbf{h}}
\def\hh{\hat{\mathbf{h}}}
\def\V{\mathbf{V}}

\def\VH{\mathbf{V}_\text{H}}

\def\VHV{\mathbf{V}_\text{A}}

\begin{document}

\maketitle

\begin{abstract}
Buried landmines and unexploded remnants of war are a constant threat for the population of many countries that have been hit by wars in the past years.
The huge amount of human lives lost due to this phenomenon has been a strong motivation for the research community toward the development of safe and robust techniques designed for landmine clearance.
Nonetheless, being able to detect and localize buried landmines with high precision in an automatic fashion is still considered a challenging task due to the many different boundary conditions that characterize this problem (e.g., several kinds of objects to detect, different soils and meteorological conditions, etc.).
In this paper, we propose a novel technique for buried object detection tailored to unexploded landmine discovery.
The proposed solution exploits a specific kind of convolutional neural network (CNN) known as autoencoder to analyze volumetric data acquired with ground penetrating radar (GPR) using different polarizations.
This method works in an anomaly detection framework, indeed we only train the autoencoder on GPR data acquired on landmine-free areas.
The system then recognizes landmines as objects that are dissimilar to the soil used during the training step.
Experiments conducted on real data show that the proposed technique requires little training and no ad-hoc data pre-processing to achieve accuracy higher than $93\%$ on challenging datasets.
\end{abstract}



\IEEEpeerreviewmaketitle

\section{Introduction}\label{sec:introduction}

In the last seventy years, landmines have been massively deployed in a huge amount of countries all over the world.
It is not possible to provide a global estimate of the total area contaminated by landmines, due to a lack of data.
However, global deaths and injuries from landmines have hit a ten-year high, with the latest available figures from 2016 showing a sharp increase on the previous year \cite{BanLandmines2017}.

The United Nations Department of Humanitarian Affairs (UNDHA) has very strict regulations in terms of civil area demining.
As a matter of fact, the $99.6\%$ of mines and unexploded ordnance must be safely removed from an area in order to consider it landmine-free \cite{imas2001}.
This is strongly different from military demining, in which case only a few paths need to be freed to enable vehicles passage, and little attention is paid to the surrounding areas.
For this reason, humanitarian demining operations are typically performed manually making humanitarian demining an extremely risky and long operation.

In order to ease demining operations and make them safer, different systems have been proposed in the literature.
In particular, landmine clearance systems generally follow a common pipeline, and different solutions have been proposed for each step of it.
These steps can be summarized as:
\begin{itemize}
\item \textit{Detection}: given an area of investigation, detecting whether possibly dangerous objects are buried. During this step, it is important to have a very low percentage of false negatives (i.e., landmines not detected), whereas it is possible to tolerate high false positive rates (i.e., non-explosive objects detected as dangerous), which will be reduced during the next step.
\item \textit{Localization}: given that a target has been detected within an area, providing more accurate localization information about the position of the buried object.
\item \textit{Recognition}: given that a buried object has been detected and localized, being able to understand whether it actually is a landmine, or it is a misclassification of the detection step (e.g., a stone, tree branches, etc.).
\end{itemize}

In this paper, we focus on the detection problem, by proposing a method based on convolutional neural networks (CNNs) applied to ground penetrating radar (GPR) data to detect the presence of buried objects not coherent with the ground under analysis.
The choice of using GPR, methodology capable of detecting even small dielectric changes in the subsurface, is determined by the fact that the vast majority of modern landmines are mainly composed of plastic materials, thus significantly reducing the efficacy of traditional metal detector surveys.

In the last few years, many methods exploiting GPR technique to detect buried objects and landmines have been proposed.
Some of these methods are based upon model-based solutions that exploit the knowledge of wave propagation laws.
As an example, in \cite{Zhou2018} an automatic data interpretation system is proposed to estimate buried pipes from GPR B-scan images.
In \cite{DellAcqua2004} GPR B-scans are packed together and then processed in a 3D domain in order to efficiently detect buried pipes.
In \cite{Liu2017} the authors devise a 3D-to-2D data conversion filter in order to apply the 2D Reverse-Time Migration to 3D GPR data.

Due to the impressive results obtained by machine learning in the last years, other detection methods exploit supervised learning strategies.
As an example, in \cite{Maas2013}, reflection hyperbolas are localized using pattern recognition techniques.
Recently, the problem of properly training machine learning algorithm for buried threat detection has been also studied \cite{Malof2018}.

Due to the development of deep learning solutions that often outperform classical methods in many fields (e.g., computer vision, pattern recognition, etc.), these kinds of techniques are being applied to landmine detection.
As an example, in \cite{Dou2017}, the authors devise a real-time method for hyperbola recognition that employs a CNN for discriminating hyperbola from non-hyperbola traces in B-scans.
In \cite{Nunez-Nieto2014} the authors compare a neural network against logistic regression algorithms in discriminating between potential targets and clutter.
Additionally, in \cite{Lameri2017, Picetti2018b}, the authors propose different CNN-based strategies for landmine detection in B-scans.

In this work we exploit a particular CNN architecture known as autoencoder to analyze GPR volumetric data acquired with multiple polarizations, building upon our recent findings reported in \cite{Picetti2018b}.
The proposed method is able to detect whether an investigated volume contains buried objects that are not coherent with the kind of soil being analyzed.
This is done exploiting an anomaly detection scheme, i.e., the autoencoder is trained only using mine-free data in order to learn how to model the soil of interest.
At deployment time, the autoencoder recognizes the presence of anomalous objects such as landmines, as they are not coherent with the training data.
The proposed solution is tested on two different datasets consisting of real GPR acquisitions.
Specifically, in two different sand pits we have buried inert landmines, along with plastics and metal objects that produce GPR scans similar to those of landmines.
Results show promising when compared with state-of-the-art methods developed for the same kind of data.

In terms of novelty, the following aspects can be pointed out:
\begin{itemize}
	\item This is the first solution jointly exploiting horizontal and vertical polarizations for landmine detection using CNNs, to the best of our knowledge.
    \item The proposed solution works considering 3D volumetric data, rather than simple 2D B-scans as done in \cite{Lameri2017, Picetti2018b}.
    \item The proposed architecture is able to work in a cross-dataset scenario, i.e., we can train the CNN on a dataset, and use it on another one with limited accuracy loss, thus making the solution robust to different environments.
\end{itemize}
The code representing a running example of the proposed solution on a part of the dataset is also available\footnote{\url{https://tinyurl.com/lndmn}}. 

The rest of the paper is structured as it follows.
Section~\ref{sec:background} reports the background concepts useful to understand the rest of the paper.
Section~\ref{sec:formulation} formally introduces the detection problem faced in this work.
Section~\ref{sec:methodology} contains all the algorithmic details of the proposed solution.
Section~\ref{sec:setup} reports information about the used experimental setup.
Section~\ref{sec:results} provides the achieved numerical results compared with state-of-the-art deep-learning solutions.
Finally, Section~\ref{sec:conclusion} concludes the paper.

\section{Background}\label{sec:background}
In this section, we report some background information useful to understand the rest of the manuscript.
We start introducing some literature on the use of machine learning for geophysical data processing.
We then introduce the main notions behind autoencoders.
We finally report relevant information about GPR polarization.

\subsection{Machine Learning for geophysics}\label{subsec:MLgeophysics}
Deep learning, and in particular convolutional neural networks (CNNs), have shown very good performance in several computer vision applications such as image classification, face recognition, pedestrian detection and handwriting recognition \cite{Goodfellow2016}.

Recently, learning-based methodologies have been increasingly explored for different applications also by the geophysical community.
As an example, an application that has been analyzed by different authors is the analysis and understanding of Earth's subsurface structures.
To this purpose, in \cite{Alregib2018} the authors propose to use machine-learning based tools.
Additionally, different techniques have been compared in \cite{Zhao2015} and \cite{Bestagini2017} for the problem of facies recognition.
Another interesting application is that of anomaly detection.
In this framework, in \cite{Smith2010}, a method based on artificial neural networks to extract relevant features is proposed.
Moreover, a wide variety of more classical image processing applications have been studied in a learning-based framework.
These involve image restoration, tomography, and inverse problems \cite{Jin2017, McCann2017, Picetti2018a, Mandelli2018b}.

Recently, machine learning techniques have gained the trend also in GPR-specific applications.
In \cite{Nunez-Nieto2014} the authors compare a neural network against logistic regression algorithms in discriminating between potential targets and clutter in B-scans obtained from multi-frequency GPRs. 
In \cite{Maas2013}, buried objects are localized through a Viola-Jones learning algorithm and a Hough Transform specifically tailored to hyperbolas.
In \cite{Dou2017}, the authors devise a real-time method for hyperbola recognition that employs a CNN for discriminating hyperbola from non-hyperbola traces in B-scans, where areas of interest have been previously determined by a clustering algorithm.
In \cite{Lameri2017} a CNN classifier is devised for detecting hyperbolas in GPR B-scans.
The system is trained on a large dataset of synthetic along with a few background-only real data and it reaches good performance in terms of detection accuracy.
Finally, an alternative solution based on a one-class approach is proposed in \cite{Picetti2018b}.

Despite the growing interest in machine-learning in this community, works exploiting modern deep learning solutions are still only a few and need to be investigated.

\subsection{Convolutional Autoencoders}\label{subsec:autoencoders}
In this section, we provide the concepts behind autoencoders to clarify the rest of the paper.
For an in-depth autoencoder review, refer to \cite{Goodfellow2016}.

A CNN is a complex computational model partially inspired by the human neural system that consists of a high number of interconnected nodes, called \textit{neurons}.
Such nodes are organized in multiple stacked layers, each one performing a simple operation on its input.
In the literature, typical neuron layers are:
\begin{itemize}
\item \textit{Fully connected}: layers performing dot product between the input and a parametric matrix.
\item \textit{Convolutional}: layers applying multi-dimensional convolution to the input.
\item \textit{Transposed Convolutional / Deconvolutional}: layers applying upsampling followed by convolution to the input.
\item \textit{Activation}: layers applying a non-linear operation (e.g., hyperbolic tangent, rectification, etc.) to the input.
\item \textit{Pooling}: layers processing the input with a moving window, and outputting a selection of the samples within each window (e.g., the maximum, minimum, average, etc.).
\end{itemize}
Many of these operations depends on numeric parameters that can be tuned based on experience, so that the model is able to learn complex functions.
This is done by minimizing a cost function depending on the output of the last layer of the network.
This minimization is achieved using back-propagation coupled with an optimization method such as gradient descent and the use of large annotated training datasets, thus enabling the network to capture patterns in the input data and automatically extract distinctive features.

To train a CNN model for a specific image classification task we need:
\begin{itemize}
\item To define the metaparameters of the CNN, i.e., the sequence of operations to be performed, the number of layers, and the shape of the filters.
\item To define a proper cost function to be minimized during the training process.
\item To prepare a dataset of training and test images, annotated with labels according to the specific tasks.
\end{itemize}

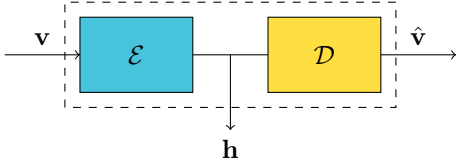
\begin{figure}
	\centering
    \begin{tikzpicture}
	\draw[->] (0,0) -- (1,0); \node[above] at (0.5,0) {$\v$};
    \draw[->] (5,0) -- (6,0); \node[above] at (5.5,0) {$\vh$};
    \draw[->] (3,0) -- (3,-1); \node[below] at (3,-1) {$\h$};
    \draw[-] (2.5,0) -- (3.5,0);
    
     \filldraw[fill=SkyBlue, draw=black] (1,-0.5) rectangle (2.5,0.5); \node at (1.75,0) {$\mathcal{E}$};

     \filldraw[fill=Goldenrod, draw=black] (3.5,-0.5) rectangle (5,0.5); \node at (4.25,0) {$\mathcal{D}$};
	
    
    \draw[dashed] (0.8,-0.7) rectangle (5.2,0.7);
\end{tikzpicture}
    \caption{Scheme of an undercomplete autoencoder. The encoder $\E$ turns the input $\v$ into its hidden representation $\h$, which is turned into $\vh$ by the decoder $\D$.}
    \label{fig:autoencoder}
\end{figure}

In this paper we exploit a specific class of CNN, the \textit{autoencoder}, that takes its name from the ability of being logically split into two separate components, as shown in Fig.~\ref{fig:autoencoder}:
\begin{itemize}
\item The \textit{encoder}, which is the operator $\E$ mapping the input $\v$ into the so called hidden representation $\h = \E(\v)$.
\item The \textit{decoder}, which is the operator $\D$ that decodes the hidden representation into an estimate of the original input $\vh = \D(\h)$.
\end{itemize}

In general, we can state that autoencoders learn how to reconstruct the input data by going to/coming from the hidden space. This is done by using as training cost function some distance metric between the input $\v$ and the output $\vh = \D(\E(\v))$.

Specifically, we make use of an \textit{undercomplete convolutional autoencoder}, i.e., a specific autoencoder characterized by a hidden representation $\h$ of reduced dimensionality with respect to the input $\v$.
By using this kind of autoencoder it is possible to estimate an almost-invertible dimensionality reduction function $\E$ directly from a representative set of training data (i.e., observations of $\v$).
In the light of this, we can interpret the hidden representation $\h = \E(\v)$ as a compact feature vector capturing salient information from $\v$.

\subsection{GPR Polarization}\label{subsec:GPR}
In this section we briefly describe the electromagnetic polarization of a GPR antenna, with a special focus on the benefits provided by combining orthogonally polarized dipole antennas in soil imaging applications.

To quantitatively characterize GPR images, it is necessary to account for the vectorial nature of electromagnetic (EM) wavefield propagation and the characteristics of the subsurface scatterers.

The use of multiple polarizations can provide key additional information, as the polarization information contained in the waves backscattered from a given target is highly related to its geometrical structure and orientation, as well as to its physical properties \cite{federico2, federico10}.

The advantages of using polarimetric radar systems for the characterization of intrinsic target properties arise from two main factors:
(i) the vector information contained in the target backscattered wave is retained (by reception diversity), and;
(ii) the entire backscattering behavior of a target can be obtained (by transmission diversity).

Therefore, radar imagery collected using different polarization combinations may provide different and complementary information \cite{federico8}.
The power of a wave scattered from an isotropic target (e.g., a sphere) is independent of the transmitter polarization.
For linear targets such as pipes and utilities, the polarization of the scattered field is independent of the transmitting polarization \cite{federico6}. 
For a general target, instead, both the power and polarization of the reflected wave vary with transmitter polarization \cite{federico7}.

Polarization is understood to have a significant impact for the identification of elongated objects and asymmetrical subsurface features thanks to their explicit polarimetric behavior \cite{federico1, federico5}.
For this reason it is largely employed as a further tool that could provide additional information to correctly reconstruct complex environments \cite{federico9}\cite{federico3}.

Wave propagation in anisotropic and heterogeneous media may also cause a change of polarization plane \cite{federico12, federico14}, due to the different propagation mechanisms encountered through the path.
Wavelet depolarization occurs to some degree for most cases of reflection \cite{federico11}.
The severity of depolarization depends on the contrast in electrical properties, incident angle, incident polarization and the orientation of the reflecting area \cite{federico15}.
The consequence is that preferential ways through which investigate targets buried in a medium with these characteristics may exist \cite{federico12, federico4}.

For this reason, in this work we exploit two different polarizations (i.e., horizontal and vertical) in order to have a better understanding of the underground.
Specifically, we combine information coming from these two orthogonal polarizations through stacking, to better capture the presence of buried objects in radar images.

\section{Problem Formulation}\label{sec:formulation}
An adequate solution to the problem posed by landmines implies that the percentage of detected mines in the area under analysis should approach $100\%$.
Furthermore, detection must be performed at the fastest possible rate, and with the lowest possible number of false negatives (i.e., mistaking a mine for a non-dangerous object).
To give a reference, the United Nations have set the detection goal at $99.6\%$, whereas the U.S. Army allows one false alarm every $1.25$ square meters.
Unfortunately, no existing landmine detection system meets these criteria.
This could be conferred to the properties of the mines themselves and the variety of environments in which they are buried, as well as to the limits and flaws in the current technology \cite{Bello2013}.

The process of landmine identification is threefold: first, the soil is analyzed in order to find the presence of buried objects; then, a more accurate analysis is performed in order to localize the targets in 3D; finally, the volume of interest is investigated with classification techniques in order to discriminate landmines from rocks and other buried objects.

In this paper we focus on the first step.
Making use of the GPR technology, the purpose of the proposed methodology is to detect whether a soil volume contains any trace of buried objects, whose dimensions and electromagnetic signature are compatible to those of landmines.

To formalize the goal of the proposed method, let us define a B-scan acquired with a GPR system as the 2D image $\V(t, x)$ whose coordinates are the reflection time $t$ and the \textit{inline} axis $x$.
By concatenating $Y$ consecutive B-scans, we sample the soil also in the third dimension, the \textit{crossline} axis $y$, thus obtaining the volume $\V(t,x,y)$.

If the $y$-th B-scan has been acquired over a buried target, we associate to it the binary label $l(y)=1$ indicating the presence of an object underground.
If it has been acquired over a target-free area, we label it with $l(y)=0$, indicating that no object traces are present.

Solving buried object detection problem under this hypothesis consists in taking the volume $\V(t,x,y)$ as input, and outputting a label $\hat{l}(y)$ (i.e., an estimate of $l(y)$) for each B-scan (i.e., $y \in \left[1, \, Y \right]$).
Correct detection occurs every time $\hat{l}(y) = l(y)$. Misclassification occurs in case $\hat{l}(y) \neq l(y)$.

Notice that this way of proceeding is challenging the algorithm performance.
As a matter of fact, in a practical scenario, detecting an object within a B-scan would be sufficient for its removal, even if the object is big enough to span multiple B-scans.
However, we consider the more challenging scenario in which all B-scans showing traces of an object must be labeled correctly.

\section{Proposed Methodology}\label{sec:methodology}
As reported in Section~\ref{subsec:autoencoders}, the autoencoder learns how to encode and decode the training data.
Once trained, if the input data are too different from that of the training set, the autoencoder fails in reconstructing it correctly, as the encoding / decoding procedure is highly suboptimal.
Therefore, the error introduced in encoded or decoded data can be used as an indicator of anomaly of the processed data with respect to the training data.
For this reason, the autoencoder can be a powerful instrument for anomaly detection \cite{Cozzolino2016, Yarlagadda2018}.

In our application scenario, it is possible to train an autoencoder to learn a hidden representation of small GPR volumes not showing any object trace.
After training, this autoencoder processes the whole data volume under analysis.
Then, an anomaly mask is computed, and a label $\hat{l}=0$ or $\hat{l}=1$ is associated to each B-scan, depending on the value of the anomaly metrics.

In the following, we describe each step of the proposed method, from data pre-processing, to network training and testing procedures.

\begin{figure}[t]
	\centering
    \begin{tikzpicture}
	\draw[->] (0,0) -- (1,0); \node[above] at (0.5,0) {$\v$};
    \draw[->] (5,0) -- (6,0); \node[above] at (5.5,0) {$\vh$};
    \draw[-] (2.5,0) -- (3.5,0);
    \draw[-] (7.5,0) --(8,0);
    \draw[->] (3,0) -- (3,-1) -- (5.3,-1); \node[above] at (3,0) {$\h$};
    \draw[->] (8,0) -- (8,-1) -- (5.7,-1); \node[above] at (8,0) {$\hh$};
    
    
     \filldraw[fill=SkyBlue, draw=black] (1,-0.5) rectangle (2.5,0.5); \node at (1.75,0) {$\mathcal{E}$};
     \filldraw[fill=Goldenrod, draw=black] (3.5,-0.5) rectangle (5,0.5); \node at (4.25,0) {$\mathcal{D}$};    
     \filldraw[fill=SkyBlue, draw=black] (6,-0.5) rectangle (7.5,0.5); \node at (6.75,0) {$\mathcal{E}$};
    
    \draw (5.5,-1) circle [radius=0.2]; \node at (5.5,-1) {$-$};
    \draw[->] (5.5, -1.2) -- (5.5, -1.7); \node[right] at (5.5, -1.7) {$e = \vert \h - \hh \vert_2$};
\end{tikzpicture}
    \caption{The proposed anomaly detection scheme. A block under analysis $\v$ is autoencoded to $\vh$ and encoded again into $\hh$. High Euclidean distance in the hidden space indicates anomaly.}
    \label{fig:method}
\end{figure}
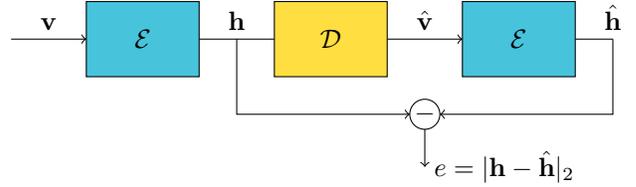

\subsection{Data Pre-processing}\label{subsec:preprocessing}
Let us define the data volume obtained with horizontal polarization as $\V_\text{H}(t, x, y)$, and the relative volume obtained with vertical polarization as $\V_\text{V}(t, x, y)$.
The first step of our algorithm consists in merging these two volumes into a single one but taking advantage of what the different polarization can detect.

To do so, we separately align each pair of A-scans acquired with horizontal and vertical polarization, i.e., $\V_\text{H}(t)$ and $\V_\text{V}(t)$.
This is done by estimating the time shift $\tau$ between them as the amount of samples corresponding to the lag that maximizes the cross-correlation between the A-scans, i.e.,
\begin{equation}
	\hat{\tau} = \arg\max_\tau \chi(\tau),
\end{equation}
where $\chi(\tau)$ is the cross-correlation between $\V_\text{H}(t)$ and $\V_\text{V}(t)$.
We then merge the two A-scans into a single one by re-alignment and averaging as
\begin{equation}
	\VHV(t) = \frac{\V_\text{H}(t) + \V_\text{V}(t - \hat{\tau})}{2}.
\end{equation}

A new data volume $\VHV(t, x, y)$ is then generated by simply concatenating all aligned A-scans $\VHV(t)$ coherently.

\subsection{Convolutional Autoencoder}\label{subsec:autoencoder}
After the data volume has been processed, we can feed it to the proposed autoencoder.
In the following, we report the strategy we propose to split the data volume into smaller blocks, train the autoencoder, and use it at test time.

\subsubsection{Block Extraction}
\begin{figure}[t]
	\centering
    \scalebox{.7}{
	\input{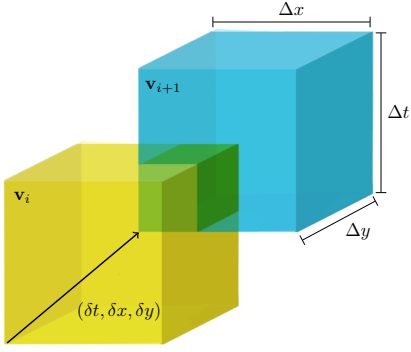}}
	\caption{Representation of the extraction of two adjacent blocks $\mathbf{v}_i$ (yellow) and $\mathbf{v}_{i+1}$ (blue) from volume $\V$. Being $\delta t < \Delta t, \delta x < \Delta x, \delta y < \Delta y$ the blocks are overlapped in the green sub-block.}
	\label{fig:blocks}
\end{figure}

Rather than processing the available data volume $\V(t,x,y)$ as a whole\footnote{Let us drop the subscript $\textrm{A}$ for the sake of notational compactness.}, we split it into a subset of $I$ regular smaller blocks $\v_i(t,x,y), \; i \in [1, I]$.
These blocks can be either overlapped or not, depending on the desired trade-off between detection robustness and computational complexity.
With reference to Figure~\ref{fig:blocks}, the $i$-th volume blocks can be defined as
\begin{equation}
  \begin{split}
      & \v_i(\tilde{t}, \tilde{x}, \tilde{y}) = \V(t, x, y), \\
      & \;\;\; \tilde{t} \in [1, \Delta t], \tilde{x} \in [1, \Delta x], \tilde{y} \in [1, \Delta y], \\
      & \;\;\; t \in [t_i, t_i + \Delta t], x \in [x_i, x_i + \Delta x], y \in [y_i, y_i + \Delta y],  \end{split}
 \end{equation}
where $\Delta t$, $\Delta x$ and $\Delta y$ are parameters that define the block size, whereas the differences $\delta t = (t_{i+1} - t_i)$,  $\delta x = (x_{i+1}-x_i)$ and  $\delta y = (y_{i+1}-y_i)$ determine blocks overlap and are known as strides, i.e., the amount of samples of the original volume between one block and the next one.
Notice that, in case of overlap (i.e., $\Delta t >\delta t$, or equivalently for $x$ and $y$), one sample of $\V$ will belong to many blocks $\v_i$. 
Moreover, in case of $\Delta y=1$, volume blocks $\v_i$ collapse into 2D B-scan patches (a suboptimal choice which will be cleared in the results presentation).

The motivation behind the choice of using small volumetric blocks is threefold:
\begin{itemize}
\item Splitting volumes into smaller blocks allows us to obtain a greater amount of data for training, considering that acquiring a B-scan is time consuming.
\item Feeding the CNN with smaller data volumes enables working with smaller and lighter architectures.
\item The system becomes independent from the whole volume size.
\end{itemize}

\begin{figure*}[t]
	\centering
    \input{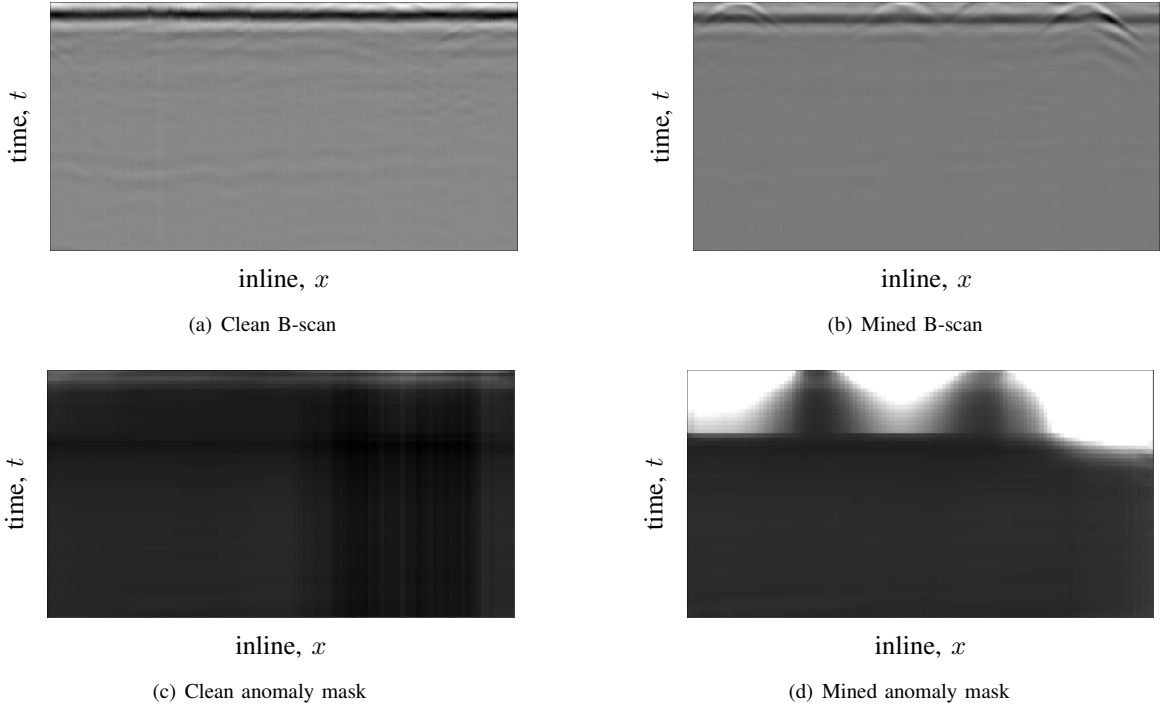}
    
    
	\caption{Example of background B-scans (landmine-free \subref{fig:bsc0_bscan} and mined \subref{fig:bsc1_bscan}) with the corresponding anomaly masks \subref{fig:bsc0_mask} and \subref{fig:bsc1_mask} normalized on the same scale of values. It is possible to see that anomalies due to hyperbola are clearly highlighted in white.}
	\label{fig:bsc}
\end{figure*}

\subsubsection{System Training}
As shown in Figure~\ref{fig:autoencoder}, the proposed architecture takes a volume $\v_i$ as input, it shrinks it to a smaller dimensionality representation $\h_i$, and it outputs a volume $\vh_i$ of the same size of the input.
To train the autoencoder, we define a training set of $I$ blocks $\v_i$, $i \in [1, I]$ extracted from $B$ adjacent B-scans associated to label $l = 0$ (i.e., the selected volume does not containing any hyperbola due to buried objects).
We then estimate the autoencoder weights by minimizing a loss function $\mathcal{L}$ defined as
\begin{equation}
	\mathcal{L} = \frac{1}{I} \sum_{i=1}^{I} \vert \v_i - \vh_i \vert_2 ^2,
	\label{eq:loss}
\end{equation}
where $\vert \cdot \vert_2$ represents the $\ell^2$-norm of a vector.
In other words, the used loss function is the mean squared error between the input volume $\v_i$ and the autoencoder output $\vh_i$ averaged over the whole training set.
Once the system has converged, we can state that the autoencoder has learnt how to correctly reconstruct background soil information.

\subsubsection{System Deployment}
When a volume of soil $\V$ is to be analyzed, we split it into a set of $I$ blocks $\v_i$, $i \in[1,I]$ covering the whole $\V$.
Figure~\ref{fig:method} explains the anomaly detection procedure.
Each block is encoded into its hidden representation $\h_i = \mathcal{E}(\v_i)$.
The hidden representation is decoded into $\vh_i = \mathcal{D}(\h_i)$, which is encoded again into $\hh_i = \mathcal{E}(\vh_i)$.
We then compare the hidden representations of both the original and autoencoded blocks by means of Euclidean distance, thus computing
\begin{equation}
	e_i = \vert \h_i - \hh_i \vert_2.
	\label{eq:distance}
\end{equation}
The distance between the hidden representations can be used as anomaly detector.
Indeed, blocks containing hyperbola traces produce a hidden representation $\hh_i$ very different from $\h_i$.
On the other hand, the two hidden representations are similar if the samples under analysis are similar to those of the training (i.e., they do not contain traces of buried objects).

\subsection{Aggregation}\label{subsec:aggregation}
For each block $\v_i$ of the input volume $\V$ we have computed the Euclidean distance $e_i$ according to \eqref{eq:distance}.
The goal of the aggregation step is to merge all obtained $e_i$ values into a volumetric anomaly mask $\mathbf{M}$ the same size $\V$, indicating which samples of $\V$ should be considered anomalies, thus buried objects.

If we consider the case of overlapped $\v_i$ blocks, a single sample of $\V$ belongs to different blocks $\v_i$.
Therefore, we have more $e_i$ values associated to a single sample of the volume $\V$.
For this reason, the volumetric anomaly mask $\mathbf{M}$ is constructed by an overlap-and-average operation.

In practice, let us define the set of indexes $i$ of blocks $\v_i$ containing the sample $\V(t,x,y)$ as
\begin{equation}
	\mathcal{I}_{t,x,y} = \left\lbrace i \; | \; \V(t, x, y) \subset \v_i \right\rbrace,
\end{equation}
where, with a slight abuse of notation, we use $\subset$ to express that the sample $\V(t, x, y)$ is contained in block $\v_i$.
The mask $\mathbf{M}$ is then defined as
\begin{equation}
	\mathbf{M}(t, x, y) = \frac{1}{|\mathcal{I}_{t,x,y}|} \sum_{i \in \mathcal{I}_{t,x,y}} \mathbf{e}_i,
\end{equation}
where $|\mathcal{I}_{t,x,y}|$ is the cardinality of the set $\mathcal{I}_{t,x,y}$.

Once the anomaly mask $\mathbf{M}$ has been obtained, in order to detect landmines in the $y$-th B-scan, we consider the maximum value in the $y$-th slice of $\mathbf{M}$ and apply the following criterion:
\begin{equation}
	\hat{l}(y) = 
	\begin{cases}
	1, & \text{if } \underset{t,x}{\max} \enskip \mathbf{M}(t,x,y)
 > \Gamma,\\
	0, & \text{otherwise},\\
	\end{cases}
    \label{eq:threshold}
\end{equation}
where $\Gamma$ is a global threshold to be selected upon a set of training B-scans at system tuning time.
In other words, if we detect a big anomaly in a B-scan, that B-scan is labeled as suspicious.

Figure~\ref{fig:bsc} shows examples of processed B-scans along with the corresponding slices of the anomaly mask. In particular, to a B-scan containing any trace of hyperbolas it correspond a uniformly low-valued anomaly slice; on the other hand, when the B-scan does contain hyperbolas, the correspondent mask clearly shows high values of the anomaly metrics.

    

    

\section{Experimental Setup}\label{sec:setup}

In this section we report the details related to all tested CNN architectures, the considered acquisition system, as well as the datasets used during our experimental campaign.

\subsection{Autoencoder Architectures}\label{subsec:architectures}
In the following, we describe the six CNN architectures we have tested to identify the best candidate solution.
Specifically, we considered three different architectures, each one working with either 2D patches, or 3D volume blocks as input.
All architectures are symmetric, as to each convolutional layer used at the encoder, corresponds a deconvolutional layer at the decoder.
The input size of each network is equal to its output size, as per autoencoder definition.
The main difference between the architectures is the hidden representations dimensionality: different architectures compress more (or less) the input data.
We tested these different autoencoder architectures to investigate the impact of this reduction.

The first architecture is $\mathcal{A}_1$, and it is composed by:
\begin{itemize}
	\item One 2D convolutional layer with 16 filters, stride 1$\times$1, size 6$\times$6.
	\item Three 2D convolutional layer with 16 filters, stride 2$\times$2, size 5$\times$5, 4$\times$4, 3$\times$3, respectively.
	\item A 2D convolutional layer with 16 filters, stride 2$\times$2, size 2$\times$2. Its output is the hidden representation.
	\item Four 2D deconvolutional layers with 16 filters, stride 2$\times$2, size 2$\times$2, 3$\times$3, 4$\times$4, 5$\times$5, respectively.
	\item One 2D deconvolutional layers with 1 filter, stride 1$\times$1, size 6$\times$6, followed by hyperbolic tangent activation.
\end{itemize}
This architecture shrinks the input by a factor 16 (e.g., a 32$\times$32 image is turned into a 64 element hidden representation).

Architecture $\mathcal{A}_2$ is the same as $\mathcal{A}_1$, but the convolutional layer returning the hidden representation is substituted by one 2D convolutional layer with 8 filters, stride 1$\times$1, size 1$\times$1.
This architecture shrinks the input by a factor 32 (e.g., a 32$\times$32 image is turned into a 32 element hidden representation).

Architecture $\mathcal{A}_3$ is the same as $\mathcal{A}_1$, but the convolutional layer returning the hidden representation is substituted by three layers:
\begin{itemize}
	\item One 2D convolutional layer with 16 filters, stride 2$\times$2, size 2$\times$2.
	\item One 2D convolutional layer with 16 filters, stride 2$\times$2, size 1$\times$1.
	\item One 2D deconvolutional layer with 16 filters, stride 2$\times$2, size 2$\times$2.
\end{itemize}
This architecture shrinks the input by a factor 64 (e.g., a 32$\times$32 image is turned into a 16 element hidden representation).

When these architectures are applied to 2D patches (i.e., blocks $\v_i$ considering the case of $\Delta y = 1$), the input has size $\Delta t \times \Delta x \times 1$.
In the 3D scenario (i.e., blocks $\v_i$ considering the case of $\Delta y \neq 1$), the input has size $\Delta t \times \Delta x \times 3$.
In other words, we stack patches from three neighboring B-scans in the third dimension, as it is typically done for the color channels in the image processing literature.

In the following, in the 2D scenario, we refer to these architectures as $\mathcal{A}_1^\text{2D}$, $\mathcal{A}_2^\text{2D}$ and $\mathcal{A}_3^\text{2D}$, respectively.
In the 3D scenario, we refer to these networks as $\mathcal{A}_1^\text{3D}$, $\mathcal{A}_2^\text{3D}$ and $\mathcal{A}_3^\text{3D}$, respectively.

All networks have been trained using Adam optimizer \cite{Kingma2015} with default parameters, until loss function stopped decreasing on a small set of validation data taken from the training set.
Network input was always normalized in range $[-1, 1]$.

\subsection{Acquisition System}\label{subsec:system}
\begin{figure}[t]
	\centering
	\subfigure[Main sensor]{\includegraphics[width=.45\columnwidth]{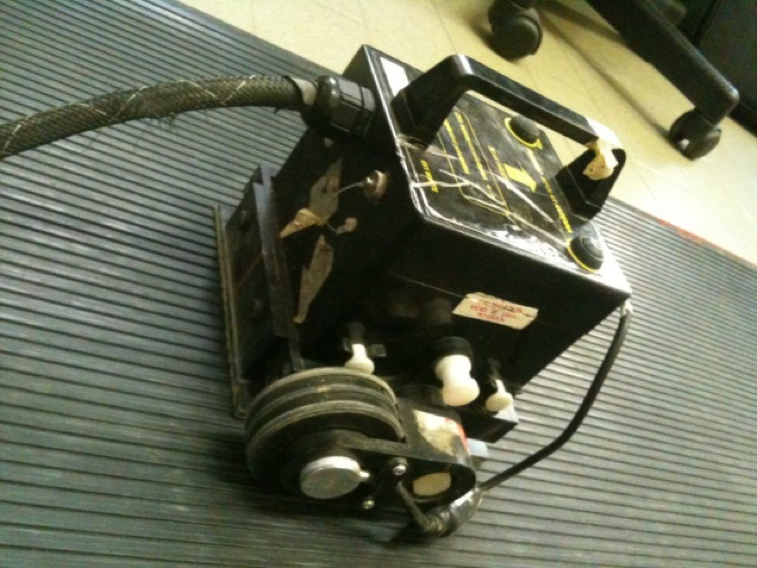}\label{fig:radar}}\hfil
    \subfigure[Dipole]{\includegraphics[width=.25\columnwidth]{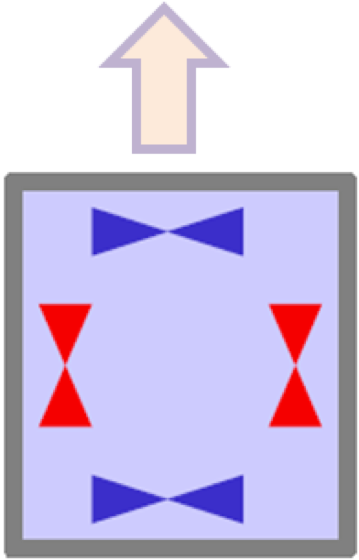}\label{fig:dipole}}
	\caption{Employed GPR system. \subref{fig:radar} Main sensor. \subref{fig:dipole} Dipoles scheme and geometry.}
	\label{fig:system}
\end{figure}

The GPR system employed for the measurements consisted of an IDS Aladdin radar (provided by IDS Georadar srl, Figure~\ref{fig:radar}), an impulse device carrying dipole antennas with a central frequency and a bandwidth of 2 GHz.
The equipment is composed by two pairs of orthogonally polarized dipole antennas, as shown in Figure~\ref{fig:dipole}, located such that the reflection center corresponds for both couples and coincides with the geometrical center of the unit.

This configuration guarantees precise matching between the parallel (i.e., vertical) and perpendicular (i.e., horizontal) orientation, in respect to the survey direction, permitting joint orthogonally polarized scans to be acquired in a single pass.
Data recording is controlled by an odometric wheel directly connected to the sensor.

Acquisitions were carried out employing a semi-mechanical device (visible in Figure~\ref{fig:radar}, \cite{federico16}), a solution which guarantees accurate data density and regularity, maintaining a precise profile spacing and ensuring a constant antenna orientation during the whole survey.

\subsection{Acquisition Campaigns}\label{subsec:campaigns}
In order to properly evaluate the proposed solution, we acquired a set of real GPR data.
The acquisitions were made in two different sand pits of width 1 meter and length 2 meters. The targets were buried at a depth of 5-15 cm, approximately.
Specifically, we focused on two different scenarios with incremental landmine detection difficulty.

\begin{figure}[t]
	\centering
	\subfigure[$\mathcal{S}_1$]{\includegraphics[width=.78\columnwidth]{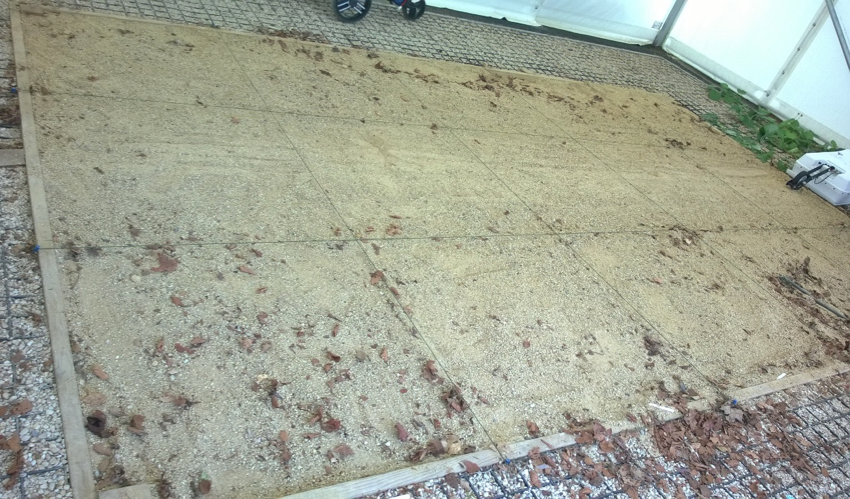}\label{fig:cranfield}}
    
    \subfigure[$\mathcal{S}_2$]{\includegraphics[width=.5\columnwidth]{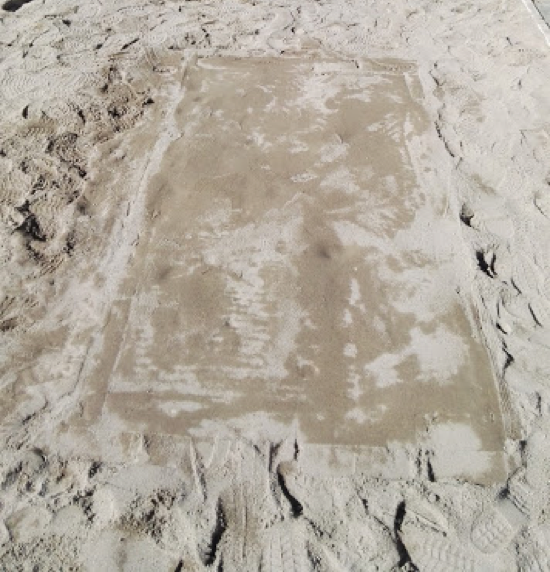}\label{fig:giuriati}}
	\caption{Environment details for datasets $\mathcal{S}_1$ and $\mathcal{S}_2$.}
	\label{fig:cranfield_giuriati}
\end{figure}

\textbf{Dataset $\mathcal{S}_1$:}
The dataset is composed by 9 different objects: 3 real landmines (SB33, PFM1, VS50), 2 printed replica of the VS50, a rubber disk, a crushed aluminum can, a metal sphere and a mortar simulant. 
The test bed is a indoor confined bay composed of several quadrants (see Figure~\ref{fig:cranfield}), and filled with a sharp sand material characterized by a very low clay content and a gritty texture, to avoid trench artifacts in the collected data.
The material was relatively homogeneous and free of clutter, with an average particle size of less than half centimeter.

Despite the environment humidity, the sand maintained a velocity of $14 \, \textrm{cm}/\textrm{ns}$ and a consequential relative dielectric constant of $4.5$.
Although a favorable propagation environment, the material is representative of several mine-affected regions of the world. 

\textbf{Dataset $\mathcal{S}_2$:}
The second scenario consists of 8 different objects: 3 real landmines (PROM1, DM11 and PMN58), a grenade, a metal sphere, an aluminum can, a plastic bottle and a plant root.
as shown in Figure~\ref{fig:giuriati}.
In this case, due to the rain falls that occurred few days before the experiment, the sand was not completely dry, providing a propagation velocity of $10 \, \textrm{cm}/\textrm{ns}$ and a resulting dielectric value of $9$.
The consequence is that the medium may not be strictly considered homogeneous, as the shallow subsurface is expected to be more dry compared to the higher moisture level of the deeper layers.

Acquisition details for the two sites are provided in Table~\ref{tab:datasets}.

\begin{table}[t]
\centering
\caption{Acquisition details for datasets $\mathcal{S}_1$ and $\mathcal{S}_2$.}
\label{tab:datasets}
\begin{tabular}{|l|c|c|}
\hline
\textbf{Parameter} & $\mathbf{\mathcal{S}_1}$ & $\mathbf{\mathcal{S}_2}$ \\
\hline \hline
Central frequency / Bandwidth & 2 GHz / 2 GHz & 2 GHz / 2 GHz \\
Propagation velocity in soil & 14 cm/ns & 10 cm/ns \\
Time sampling & 0.0117 ns & 0.0117 ns \\
Inline sampling & 0.4 cm & 0.4 cm \\
Crossline sampling & 0.8 cm & 1.6 cm \\
Time window & 15 ns & 20 ns \\
Acquired B-scans & 114 & 66 \\
\hline
\end{tabular}
\end{table}

\section{Results}\label{sec:results}
In this section, we present the experimental results obtained with the proposed methodology.
First, we introduce the used detection metric.
Then, we describe in detail all the performed tests along with their numerical performances.
Finally we provide a few comments about computational resources.
The code as well as part of the dataset are available online\footnote{\url{https://tinyurl.com/lndmn}}.

\subsection{Evaluation Metric}\label{subsec:metrics}
The proposed classification method is based on the threshold $\Gamma$ defined in \eqref{eq:threshold}.
We therefore evaluated our technique by means of receiver operating characteristic (ROC) curves.
A ROC curve represents the probability of correct detection (i.e., correctly finding a threat) and probability of false detection (i.e., detecting objects in clear areas) by spanning all possible values of the threshold $\Gamma$.
This means that each working point of a ROC curve is determined by a specific $\Gamma$ value. 
As compact measure of ROC goodness we selected the area under the curve (AUC).
This measure ranges between 0 (i.e., estimated labels are inverted with respect to the true ones) and 1 (i.e., perfect result), passing through $0.5$ (i.e., random guess).

\begin{figure}[t]
	\centering
	\scalebox{0.8}{
    \pgfplotstableread{./figures/mse_mask_max.txt} \CURVES
\small
\begin{tikzpicture}
	\begin{axis}[
    			xmin=0, xmax=109, ymin=0, ymax=4,
    			y label style={at={(axis description cs:-0.05,.5)}},
				xlabel={$y$},
                ylabel={$\underset{t,x}{\max} \enskip \mathbf{M}(t,x,y)$},
				grid=major,
                legend pos=north west,
                legend cell align=left
                ]
                
      	\addplot[color=ProcessBlue, ultra thick] table[x=0,y=1] from \CURVES;
        \addlegendentry{$\VH$, 2D};
        
        \addplot[color=OliveGreen, ultra thick] table[x=0,y=2] from \CURVES;
        \addlegendentry{$\VHV$, 2D};
        
        \addplot[color=RedOrange, ultra thick, yscale=10] table[x=0,y=3] from \CURVES;
        \addlegendentry{$\VHV$, 3D (10x)};
        
        \addplot[color=black, thick, yscale=1] table[x=0,y=4] from \CURVES;
        \addlegendentry{$l(y)$};
        
	\end{axis}
\end{tikzpicture}}
    \caption{Examples of maximum anomaly values (colored) related to architectures $\mathcal{A}_3$ applied to dataset $\mathcal{S}_1$. These curves are thresholded and  compared with the ground truth (black) in order to obtain the ROC curves.}
    \label{fig:MSE_vs_gt}
\end{figure}
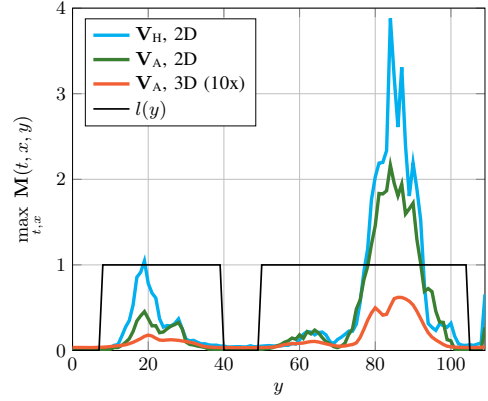

To better understand how ROC curves work, let us consider Figure~\ref{fig:MSE_vs_gt}.
This figure depicts the maximum values of the Euclidean distance $e_i$ for each B-scan (i.e., $\underset{t,x}{\max} \enskip \mathbf{M}(t,x,y)$) using three different network configurations on dataset $\mathcal{S}_1$.
This figure also reports the ground truth for each B-scan, i.e., the label curve $l(y)$ that assumes value 1 if the $y$-th B-scan contains a threat.
Each point of a ROC curve is obtained by comparing the ground truth with a binarized version of $\underset{t,x}{\max} \enskip \mathbf{M}(t,x,y)$ using threshold $\Gamma$.

In order to understand results obtained through the ROC curves, an additional comment is needed.
In a real scenario, it is common that traces generated by a single landmine can be observed in multiple B-scans.
The way we have decided to evaluate our system, we consider a mis-detection every time we miss a single B-scan, therefore all presented results can than be considered as a lower bound on the performance of the proposed method.

\subsection{Detection Performance}
In order to evaluate the detection performance of the proposed system, we performed a set of tests aiming at validating a specific choice we made.
In the following, we discuss all of these tests.

\textbf{Block size and stride:}
The first test aims at evaluating the impact of the block size and stride.
To this purpose, we worked in the 2D scenario.
We trained the networks with the first $N=5$ background B-scans of dataset $\mathcal{S}_1$ using only the horizontal polarization, and tested it on the remaining volume of data.

\begin{figure}[t]
	\centering
    \subfigure[$\mathcal{S}_1$ B-scan]{
        \begin{tikzpicture}
            \node (img) {\includegraphics[width=.7\columnwidth]{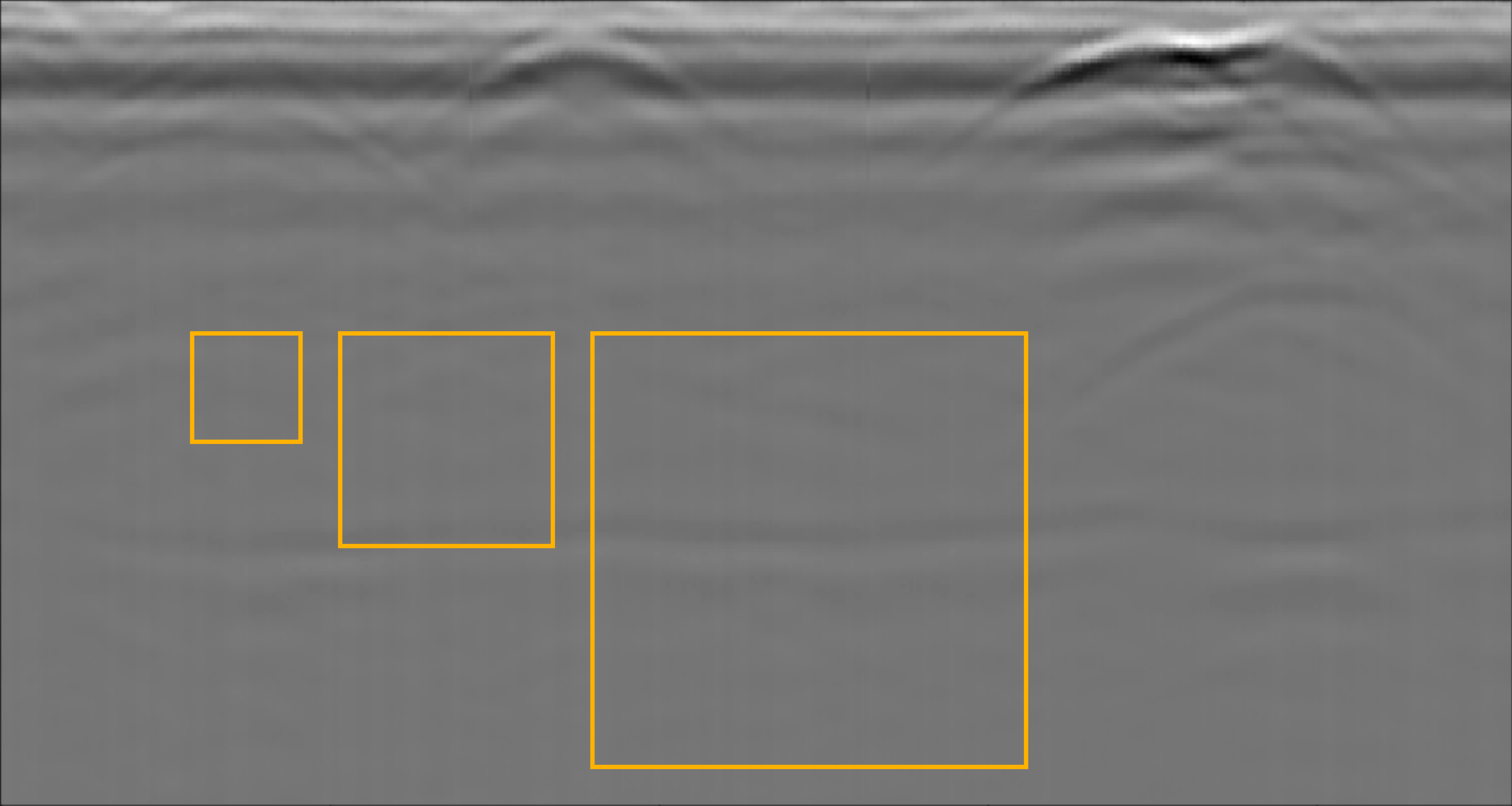}};
            \node[below=of img, node distance=0cm, yshift=1cm] {inline, $x$};
            \node[left=of img, node distance=0cm, rotate=90, anchor=center, yshift=-.7cm] {time, $t$};
        \end{tikzpicture}
        \label{fig:bscan_example_1}
    }

    \subfigure[$\mathcal{S}_2$ B-scan]{
        \begin{tikzpicture}
            \node (img) {\includegraphics[width=.7\columnwidth]{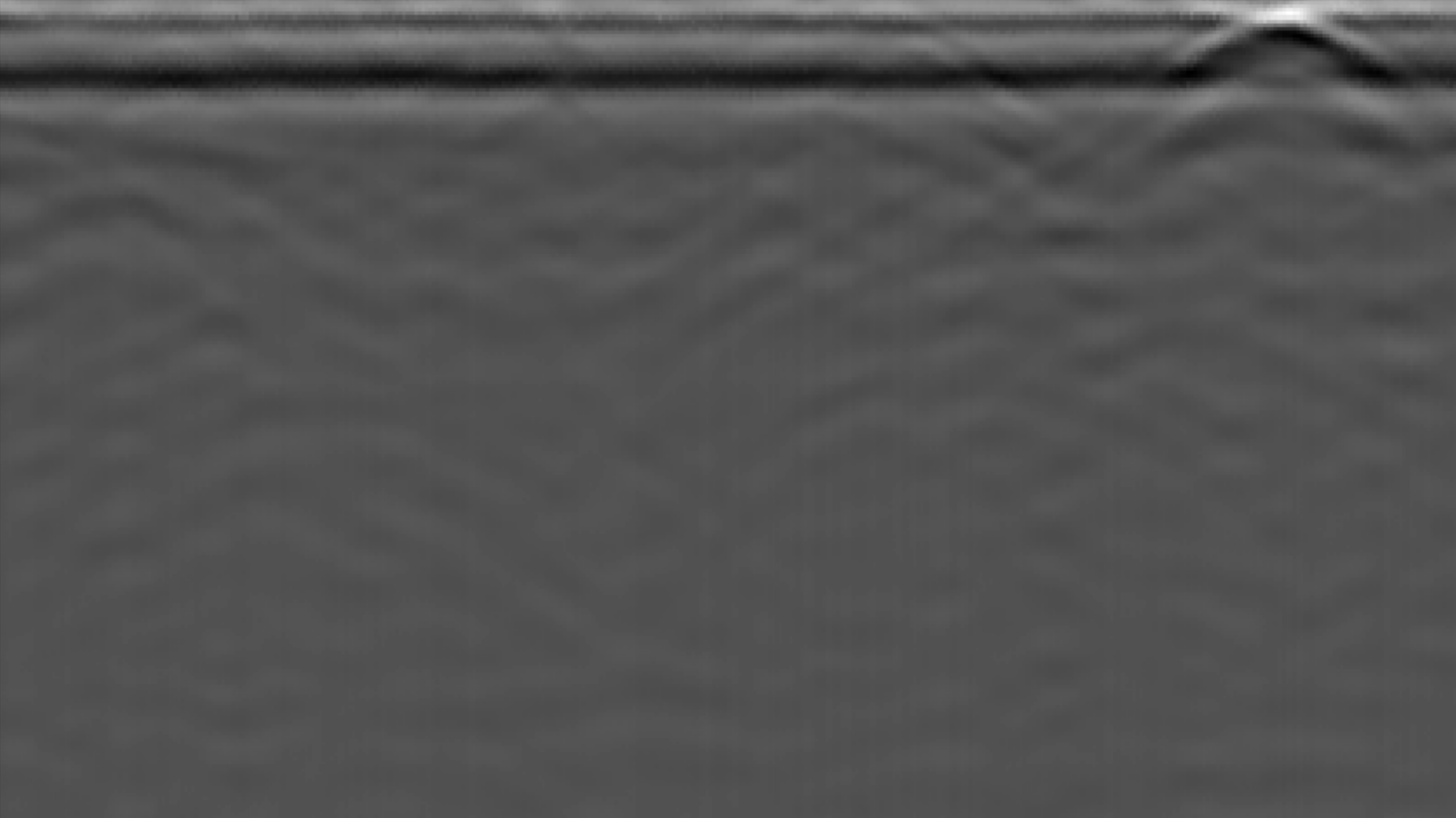}};
            \node[below=of img, node distance=0cm, yshift=1cm] {inline, $x$};
            \node[left=of img, node distance=0cm, rotate=90, anchor=center, yshift=-.7cm] {time, $t$};
        \end{tikzpicture}
        \label{fig:bscan_example_2}
    }
    \caption{Examples of B-scans taken from $\mathcal{S}_1$ \subref{fig:bscan_example_1} and $\mathcal{S}_2$ \subref{fig:bscan_example_2}. In yellow, three different squared patches of 32, 64, and 128 samples respectively.}
    \label{fig:bscan_example}
\end{figure}
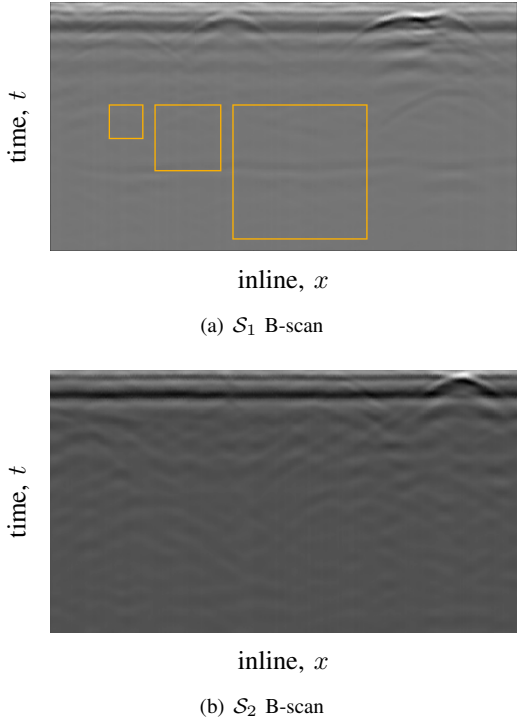

We choose arbitrary to work with squared blocks, varying the sizes $\Delta t$ and $\Delta x$ between $32$, $64$, and $128$ samples.
As the block stride is concerned, we fix $\delta t = \delta x$ and test $4$, $16$, and $32$ samples.
From Figure~\ref{fig:bscan_example}, it is possible to observe the block sizes with respect to the input data.

Table~\ref{tab:patch_properties} reports the results in terms of AUC values.
Concerning the stride, we notice that having a small stride means that blocks are highly overlapped.
Therefore, more blocks are extracted from the data volume.
This results in a smoother and more detailed anomaly mask, thus making a small stride preferable. 
Regarding the block size, $128$ samples are an oversized dimension, resulting in mediocre outcomes.
Conversely, $32$ and $64$ have almost the same performance.

In the rest of our experiment, we set $\Delta t = \Delta x = 64$ samples for two reasons:
\begin{itemize}
	\item It contains almost the whole hyperbola as suggested in \cite{Reichman2017}.
	\item It involves less memory usage in terms of CNN training and test (i.e., less patches to analyze without significant loss in accuracy).
\end{itemize}
In terms of stride, we select $\delta t = \delta x = 4$ as it provides smoother anomaly masks overall, still granting accurate results.

\begin{table}[t]
    \caption{Impact of different block sizes and strides in the 2D scenario in terms of AUC.}
    \label{tab:patch_properties}
	\centering
    \def\arraystretch{1.5}
    \begin{tabular}{|l|l|c|c|c|}
        \hline
        \multicolumn{2}{|c|}{Parameters} & $\mathcal{A}_1^\text{2D}$ & $\mathcal{A}_2^\text{2D}$ & $\mathcal{A}_3^\text{2D}$ 			\\ \hline
        																																	\hline
        \multicolumn{1}{|c|}{\multirow{3}{*}{$\Delta t = \Delta x = 32$}} 	& $\delta t = \delta x = 4$ 	& 0.9335 & 0.9470 & 0.9400	\\ \cline{2-5} 
        \multicolumn{1}{|c|}{}                          					& $\delta t = \delta x = 16$ 	& 0.9480 & 0.9455 & 0.9340	\\ \cline{2-5} 
        \multicolumn{1}{|c|}{}                          					& $\delta t = \delta x = 32$ 	& 0.9255 & 0.9285 & 0.9205	\\ \hline
        					\multirow{3}{*}{$\Delta t = \Delta x = 64$}   	& $\delta t = \delta x = 4$ 	& 0.9375 & 0.9360 & 0.9115	\\ \cline{2-5} 
        																	& $\delta t = \delta x = 16$	& 0.9475 & 0.9320 & 0.9550	\\ \cline{2-5} 
       																		& $\delta t = \delta x = 32$ 	& 0.9385 & 0.9130 & 0.8816	\\ \hline
                                            $\Delta t = \Delta x = 128$     & $\delta t = \delta x = 16$ 	& 0.9100 & 0.8646 & 0.8701	\\ \hline   
    \end{tabular}
\end{table}

\textbf{Training set size:}
Having selected the patch size $\Delta t = \Delta x=64$ and stride $\delta t = \delta x=4$, we analyze the impact of the number of B-scans needed to train the architectures.
For this test, again we consider 2D data from $\mathcal{S}_1$ using only the horizontal polarization.

Table~\ref{tab:training_bscans} reports the results of this scenario.
Notice that the system is very robust with respect to this parameter, as there are no significant accuracy losses while changing $N$.
However, as there is no reason for not using training data when available, we fix $N=5$ hereinafter.
Moreover, this enables to fairly compare 2D and 3D scenarios, as more than one B-scan is needed to generate volumetric data.
\begin{table}[t]
	\caption{Impact of different numbers of B-scans on which the architectures are trained in terms of AUC.}
    \label{tab:training_bscans}
	\centering
    \def\arraystretch{1.5}
    \begin{tabular}{|l|c|c|c|}
        \hline
        Training b-scans & $\mathcal{A}_1^\text{2D}$ & $\mathcal{A}_2^\text{2D}$ & $\mathcal{A}_3^\text{2D}$ \\
        \hline \hline
        $N = 1$ & 0.9468 & 0.9344 & 0.9247 \\ \hline
        $N = 3$ & 0.8956 & 0.9329 & 0.9195 \\ \hline
		$N = 5$ & 0.9320 & 0.9360 & 0.9230 \\ \hline
        \end{tabular}  
\end{table}

\textbf{Different polarizations:}
The experiment aims at demonstrating the improvements introduced by exploiting both the horizontal and vertical GPR polarizations, as described in section \ref{subsec:preprocessing}.
This is done by comparing results obtaining using only horizontal polarization, or both.

Table~\ref{tab:preprocessing} shows the impact of the preprocessed dataset using all the three 2D architectures on $\mathcal{S}_1$.
Notice that architecture $\mathcal{A}_3^\text{2D}$ greatly outperforms all the others, showing an AUC value of 96.5\%.
This confirms the importance of using both polarizations.
\begin{table}[t]
	\caption{Impact of the pre-processing applied to GPR polarizations in terms of AUC.}
    \label{tab:preprocessing}
	\centering
    \def\arraystretch{1.5}
    \begin{tabular}{|l|c|c|c|}
        \hline
        GPR Polarization & $\mathcal{A}_1^\text{2D}$ & $\mathcal{A}_2^\text{2D}$ & $\mathcal{A}_3^\text{2D}$ \\
        \hline \hline
        $\VH$ & 0.9320 & 0.9360 & 0.9105  \\ \hline
		$\VHV$ & 0.9170 & 0.9240 & \textbf{0.9650} \\ \hline
        \end{tabular}  
\end{table}

\textbf{Volumetric data:}
The objective of the following experiment is to compare the 2D architectures against their 3D-extended versions.
Results have been obtained on dataset $\mathcal{S}_1$.

Table~\ref{tab:architecture3D} reports these results.
For all the architectures, adding the third dimension allows to gain from 2 to 5 percentage points.
In particular, architecture $\mathcal{A}_3^\text{3D}$ obtains 98\% of AUC.
This result is well expected, as the exploitation of the high interdependence between adjacent B-scans could only help in terms of detection.
\begin{table}[t]
	\caption{Comparison of the detection performance of both the 2D and 3D architectures on pre-processed data from $\mathcal{S}_1$.}
    \label{tab:architecture3D}
	\centering
    \def\arraystretch{1.5}
    \begin{tabular}{|c|c|}
        \hline
        Architecture & AUC \\
        \hline \hline
        $\mathcal{A}_1^\text{2D}$ & 0.9170 \\ \hline
        $\mathcal{A}_2^\text{2D}$ & 0.9240 \\ \hline
        $\mathcal{A}_3^\text{2D}$ & \textbf{0.9650} \\ \hline
		\hline
        $\mathcal{A}_1^\text{3D}$ & 0.9555 \\ \hline
        $\mathcal{A}_2^\text{3D}$ & 0.9750 \\ \hline
        $\mathcal{A}_3^\text{3D}$ & \textbf{0.9755} \\ \hline
    \end{tabular}
\end{table}

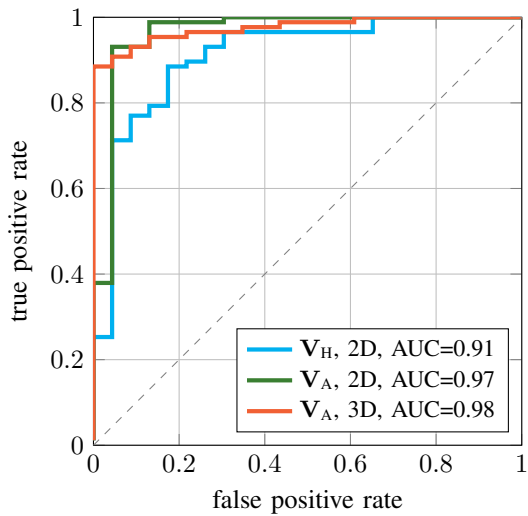
\begin{figure}[t]
	\centering
    \pgfplotstableread{./figures/roc_HH.txt} \HH
\pgfplotstableread{./figures/roc_HHVV.txt} \HHVV
\pgfplotstableread{./figures/roc_HHVV3D.txt} \HHVVD

\begin{tikzpicture}
    \begin{axis}[
    xmin=-0.0, xmax=1, ymin=0, ymax=1.0,
    xlabel=false positive rate,
    ylabel=true positive rate,
    unit vector ratio*=1 1 1,
    grid=major,
    legend pos=south east,
    legend style={font=\small}]
        \addplot[color=ProcessBlue, ultra thick] table \HH;
        \addlegendentry{$\V_\text{H}$, 2D, AUC=0.91};
        
        \addplot[color=OliveGreen, ultra thick] table \HHVV;
        \addlegendentry{$\VHV$, 2D, AUC=0.97};
        
        \addplot[color=RedOrange, ultra thick] table \HHVVD;
        \addlegendentry{$\VHV$, 3D, AUC=0.98};
        
        \pgfplotsset{
        	after end axis/.code={
            	\draw[-, color=gray, dashed](axis cs:0,0)--(axis cs:1,1);
            }
        }
    \end{axis}
\end{tikzpicture}
    \caption{ROC curves for three setups on dataset $\mathcal{S}_1$: blue, $\mathcal{A}_3^\text{2D}$ applied on horizontal polarization data; green, $\mathcal{A}_3^\text{2D}$ applied on preprocessed data; red, $\mathcal{A}_3^\text{3D}$ applied on preprocessed data.}
    \label{fig:roc}
\end{figure}

Figure~\ref{fig:roc} plots the ROC curves which shows the diagnostic ability of the binary classifier while varying the threshold $\Gamma$ applied to the anomaly detection metrics, as described in section \ref{subsec:metrics}.
In particular, we choose to show the improvements in terms of detection ability due to the adoption of multi-polarization and volumetric data.
It is worth noticing also that the 3D extension on multi-polarization data (the red line) can achieve a true positive rate of almost 90\% at a false positive rate of 0\%.
If considering buried objects rather than B-scans, the system is able to detect threats without skipping any buried objects.

These results confirm the improvement provided by using volumetric data.

\textbf{Cross-dataset:}
In computer vision and machine learning literatures, CNNs should not be extremely overfitted to training data.
Instead, the architectures should provides good performance also on datasets which have not been used for training.
It is indeed a common practice to train an architecture in order to be as general as possible, and then fine-tuning it depending on the specific target data \cite{Reichman2017b, Malof2018}.

In our scenario, we trained the best proposed architecture (i.e., $\mathcal{A}_3^\text{2D}$) on dataset $\mathcal{S}_1$, and then tested it on $\mathcal{S}_2$, and vice-versa.
Table~\ref{tab:cross-test} reports the results in terms of AUC value, which stays strictly above 93\% when the same dataset is used for training and test, and above 81\% in cross-dataset scenarios.
It can be noted that, the proposed method is robust against cross-training, thus the system is not strongly conditioned by the kind of soil used during training.

These cross-dataset results suggest that our architecture could be pre-trained on a generic dataset and then refined and deployed on a specific soil by acquiring a few background B-scans on a safe area.

\begin{table}[t]
	\caption{Results of cross-tests between $\mathcal{S}_1$ and $\mathcal{S}_2$ with preprocessed data and architecture $\mathcal{A}_3^\text{3D}$.}
    \label{tab:cross-test}
	\centering
    \def\arraystretch{1.5}
    \begin{tabular}[m]{lccc}
        & & \multicolumn{2}{c}{TEST} \\ \cline{3-4} 
        & \multicolumn{1}{c|}{} & \multicolumn{1}{c|}{$\mathcal{S}_1$} & \multicolumn{1}{c|}{$\mathcal{S}_2$} \\ \cline{2-4}
        \multicolumn{1}{c|}{\multirow{2}{*}{\begin{sideways}TRAIN\end{sideways}}} & \multicolumn{1}{r|}{$\mathcal{S}_1$} & \multicolumn{1}{c|}{0.9755} & \multicolumn{1}{c|}{0.8095} \\ \cline{2-4} 
        \multicolumn{1}{c|}{} & \multicolumn{1}{r|}{$\mathcal{S}_2$}  & \multicolumn{1}{c|}{0.8426} & \multicolumn{1}{c|}{0.9338} \\ \cline{2-4} 
        \end{tabular}
\end{table}

\textbf{Comparison against state-of-the-art:}
From the experimental campaign just presented, we conclude that the best proposed configuration consists of architecture $\mathcal{A}_3^\text{3D}$, applied to pre-processed volumetric data $\mathbf{V}_\text{A}$, split into  3D blocks characterized by $(\Delta t, \; \Delta x, \; \Delta y) = (64, \; 64, \; 3)$ and $(\delta t, \; \delta x, \; \delta y) = (4, \; 4, \; 1)$.
This justify both the use of multiple polarization and volumetric data.

In order to compare the proposed solution against other methods proposed in the literature, we consider two recently proposed CNN-based algorithms \cite{Lameri2017, Picetti2018b}.
The former exploits a two-class CNN trained jointly on synthetic and real data.
The latter is a simpler version of the autoencoder-based algorithm proposed in this paper, which does not take into account neither multi-polarization, nor volumetric data.
Both algorithms have been trained as explained in the original papers, using their best parameters and the same training data.

\begin{figure}
	\centering
    \subfigure[$\mathcal{S}_1$]{\pgfplotstableread{./figures/roc_EUSIPCO_5_9659.txt} \EUSIPCO
\pgfplotstableread{./figures/roc_TSP_9550.txt} \TSP
\pgfplotstableread{./figures/roc_HHVV3D.txt} \NEW

\begin{tikzpicture}
    \begin{axis}[
    xmin=-0.0, xmax=1, ymin=0, ymax=1.0,
    xlabel=false positive rate,
    ylabel=true positive rate,
    unit vector ratio*=1 1 1,
    grid=major,
    legend pos=south east,
    legend cell align=right,
    legend style={font=\small}]
       	\addplot[color=ProcessBlue, ultra thick, dashed] table \EUSIPCO;
        \addlegendentry{\cite{Lameri2017}, AUC=0.97};
        
        \addplot[color=OliveGreen, ultra thick, dashed] table \TSP;
        \addlegendentry{\cite{Picetti2018b}, AUC=0.95};
        
        \addplot[color=RedOrange, ultra thick] table \NEW;
        \addlegendentry{Proposed, AUC=0.98};
        
        \pgfplotsset{
        	after end axis/.code={
            	\draw[-, color=gray, dashed](axis cs:0,0)--(axis cs:1,1);
            }
        }
    \end{axis}
\end{tikzpicture}\label{fig:roc_sota_1}}
    \subfigure[$\mathcal{S}_2$]{\pgfplotstableread{./figures/roc_eusipco_giuriati_5bsc_0.595238.txt} \EUSIPCO
\pgfplotstableread{./figures/roc_tsp_giuriati_0.858289.txt} \TSP
\pgfplotstableread{./figures/roc_giuriati_0.933862.txt} \NEW

\begin{tikzpicture}
    \begin{axis}[name=roc,
    xmin=-0.0, xmax=1, ymin=0, ymax=1.0,
    xlabel=false positive rate,
    ylabel=true positive rate,
    unit vector ratio*=1 1 1,
    grid=major,
    legend pos=south east,
    legend cell align=right,
    legend style={font=\small}]
       	\addplot[color=ProcessBlue, ultra thick, dashed] table \EUSIPCO; \label{eusipco}
        \addlegendentry{\cite{Lameri2017}, AUC=0.59};
        
        \addplot[color=OliveGreen, ultra thick, dashed] table \TSP; \label{tsp}
        \addlegendentry{\cite{Picetti2018b}, AUC=0.86};
        
        \addplot[color=RedOrange, ultra thick] table \NEW; \label{new}
        \addlegendentry{Proposed, AUC=0.93};
        
        \pgfplotsset{
        	after end axis/.code={
            	\draw[-, color=gray, dashed](axis cs:0,0)--(axis cs:1,1);
            }
        }
    \end{axis}
\end{tikzpicture}\label{fig:roc_sota_2}}
    \caption{ROC curves of the proposed methodology compared with two CNN-based solutions on dataset $\mathcal{S}_1$ \subref{fig:roc_sota_1} and $\mathcal{S}_2$ \subref{fig:roc_sota_2}. Baseline state-of-the-art solutions are reported with dashed lines.}
    \label{fig:roc_sota}
\end{figure}
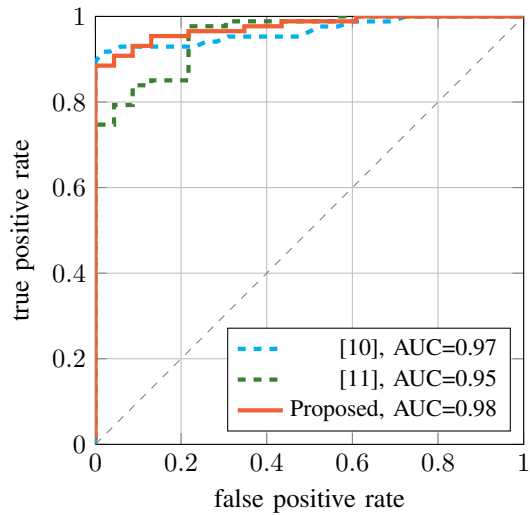
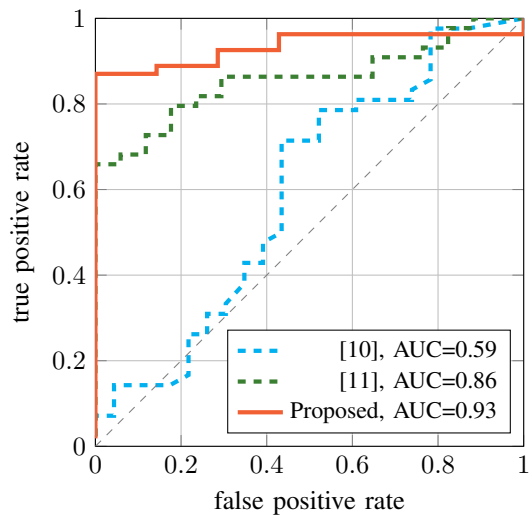

Figure~\ref{fig:roc_sota} shows the ROC curves obtained by testing the three methods (i.e., our, \cite{Lameri2017} and \cite{Picetti2018b}) on both datasets (i.e., $\mathcal{S}_1$ and $\mathcal{S}_2$).
On dataset $\mathcal{S}_1$, all methods perform very well, and the proposed one outperforms the others by only 1 to 3 percentage points in terms of AUC.
This dataset is somehow easier to deal with, as hyperbola traces are more pronounced, and the background is not particularly noisy.
Conversely, on dataset $\mathcal{S}_2$, the proposed method greatly outperforms the other state-of-the-art solutions.
As a matter of fact, $\mathcal{S}_2$ is a more challenging and realistic dataset.
Despite different kinds of synthetically generated images have been used to train the method in \cite{Lameri2017}, it performs poorly under noisy conditions.

\begin{figure}[t]
	\centering
	\scalebox{.85}{
    \pgfplotstableread{./figures/val_losses.txt} \HISTORY

\begin{tikzpicture}
    \begin{axis}[
    xmin=0, xmax=30,
    y label style={at={(axis description cs:.05,.5)}},
    xlabel=training epoch,
    ylabel=$\mathcal{L}$,
    grid=major,
    width=\columnwidth,
    height=15em,
    legend pos=north east,
    legend style={font=\small},
    ]
        \addplot[color=ProcessBlue, ultra thick] table[x=0,y=1] \HISTORY;
        \addlegendentry{$\V_\text{H}$, 2D};
        
        \addplot[color=OliveGreen, ultra thick] table[x=0,y=2] \HISTORY;;
        \addlegendentry{$\VHV$, 2D};
        
        \addplot[color=RedOrange, ultra thick] table[x=0,y=3] \HISTORY; 
        \addlegendentry{$\VHV$, 3D};
    \end{axis}
\end{tikzpicture}}
    \caption{Examples of validation loss function during the training stage. The convergence is reached in a few iterations.}
    \label{fig:val_losses}
\end{figure}
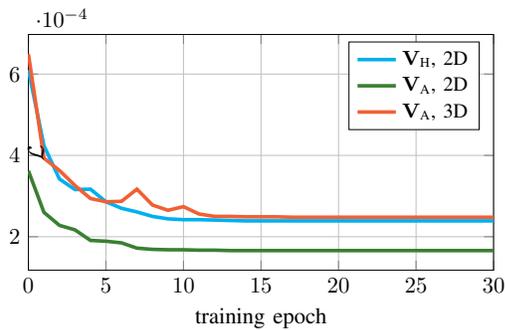

\textbf{Computational resources:}
Let us briefly comment on the effort required by the proposed workflow.
Figure~\ref{fig:val_losses} shows the curves of the loss values \eqref{eq:loss} with respect to the training epoch (i.e., the number of iterations on the whole training dataset).
In particular we can state that the proposed methodology converges reasonably fast, as after 12 epochs, the network does not improve over validation anymore.
It is therefore possible to retrain the network for each specific kind of soil just before deployment time.
As a matter of fact, all tests were performed on a workstation equipped with a Nvidia Titan X GPU reaching convergence in a few minutes.

\section{Conclusion and Future Works}\label{sec:conclusion}
In this paper, a novel technique for landmine detection by means of multi-polarization GPR acquisitions has been introduced.
The proposed system (i.e., architecture $\mathcal{A}_2^\text{3D}$ trained on multiple polarization $\V_\textrm{A}$ data) is able to detect buried threats in a volumetric dataset with an AUC of 98\%.
In particular, it reaches a detection probability of 90\% also when the desired false positive rate is zero (i.e., we do not detect any non-mine object).
A special convolutional neural network, the \textit{autoencoder}, is employed as anomaly detector.
This is trained to reconstruct background soil only, i.e., without any hyperbola whose electro-magnetic signature is similar to those of landmines.
The training dataset can be very small (i.e., just a few B-scans) and the convergence of the training is reached after a few epochs (i.e., a few minutes on a modern GPU).
Moreover, the system has shown promising performance also in a cross-dataset scenario (i.e., it has been trained on a dataset and then deployed on another dataset).

This work can be considered as an additional step toward automated humanitarian demining systems.
In this scenario, we are currently improving the proposed methodology for counting and localizing the threats.
Once geometric information about the buried threats is available, a classifier could be used to discriminate the landmines from inert buried objects.



\ifCLASSOPTIONcaptionsoff
  \newpage
\fi

\bibliographystyle{IEEEtran}
\bibliography{biblio}


\begin{IEEEbiography}[{\includegraphics[width=1in,height=1.25in,clip,keepaspectratio]{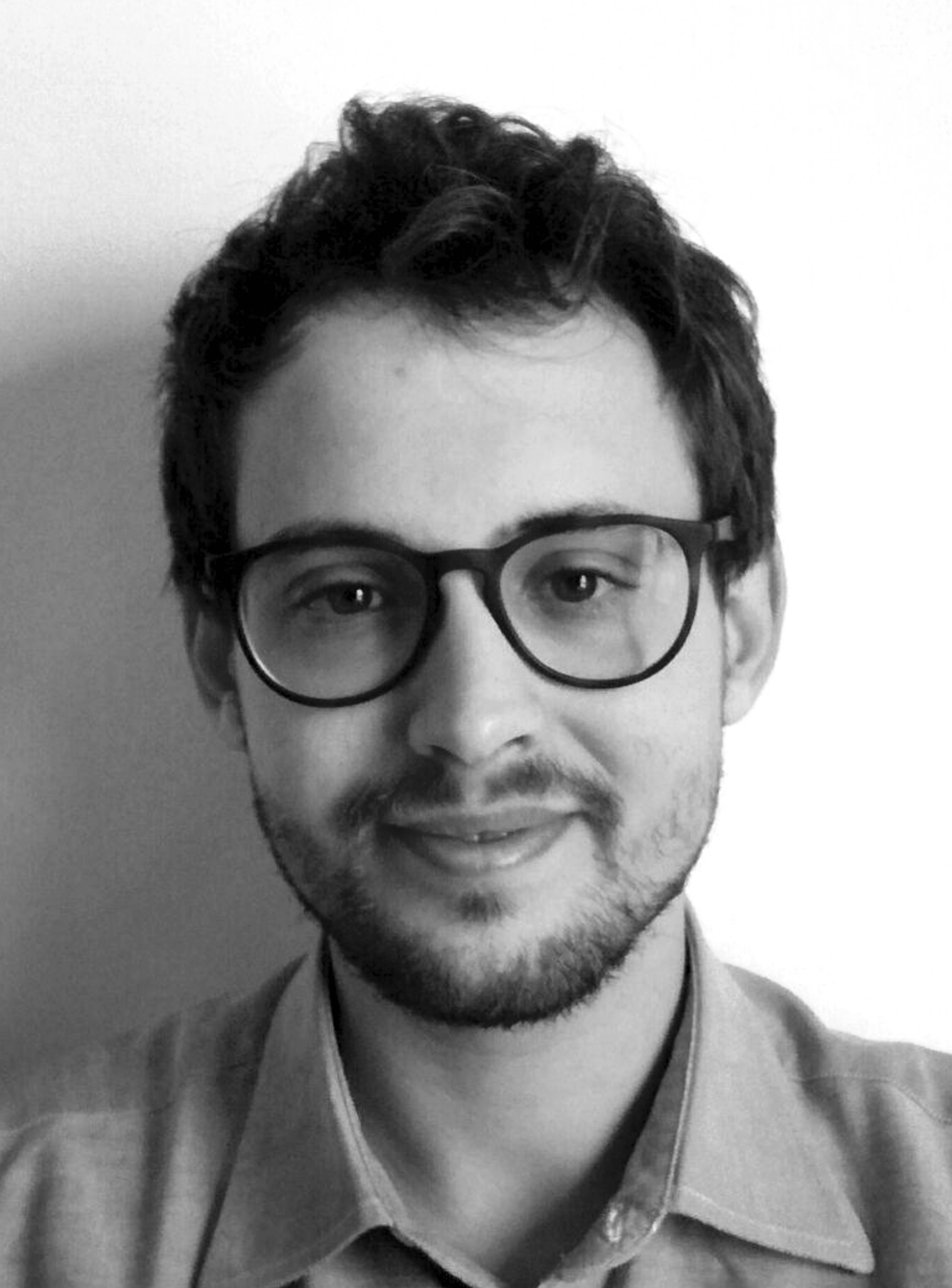}}]{Francesco~Picetti}
(S'18) was born in Milan, Italy, in 1992.
He received the B.Sc. degree in Electronic Engineering and M.Sc. degree in Computer Science and Engineering from the Politecnico di Milano, Italy, in 2015 and 2017, respectively.
He is currently a Ph.D. candidate in Information Technology working at the Image and Sound Processing Group, Politecnico di Milano.
His research interests focus on signal processing techniques for geophysical imaging.
\end{IEEEbiography}

\begin{IEEEbiography}[{\includegraphics[width=1in,height=1.25in,clip,keepaspectratio]{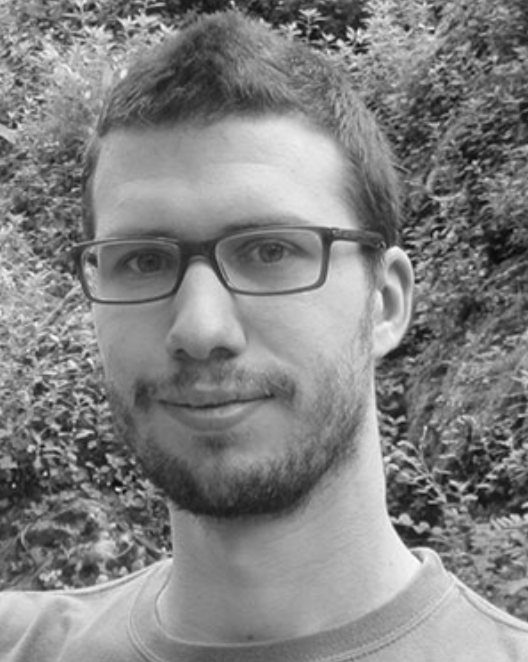}}]{Paolo~Bestagini}
(M'11) was born in Novara, Italy, on February 22, 1986.
He received the M.Sc. degree in Telecommunications Engineering and the Ph.D. degree in Information Technology from the Politecnico di Milano, Italy, in 2010 and 2014, respectively.
He is currently an Assistant Professor at the Image and Sound Processing Group, Politecnico di Milano.
His research interests focus on multimedia forensics and acoustic signal processing for microphone arrays.
He is an elected member of the IEEE Information Forensics and Security Technical Committee, and a co-organizer of the IEEE Signal Processing Cup 2018.
\end{IEEEbiography}

\begin{IEEEbiography}[{\includegraphics[width=1in,height=1.25in,clip,keepaspectratio]{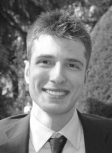}}]{Federico~Lombardi}
(GSM'17) was born in Piacenza, Italy on August 18, 1987. He received the B.Sc. and the M.Sc. degree in Telecommunications Engineering from the Politecnico di Milano, Italy, in 2009 and 2011, respectively. He joined the Radar research group at University College of London, United Kingdom, in 2015 as a Ph.D. candidate and he is currently defending his thesis on the topic "Novel Radar Techniques for Humanitarian Deminig". Since July, he is a Research Fellow at the Department of Civil and Environmental Engineering working on Ground Penetrating Radar investigation for soil characterisation. His research interests focus on non destructive testing of structures and near surface geophysical investigation. He is a nominated member of the IEEE AESS Board of Governors where he serves as the Graduate Student Representative and a member of the SEG Italian Section.
\end{IEEEbiography}


\begin{IEEEbiography}[{\includegraphics[width=1in,height=1.25in,clip,keepaspectratio]{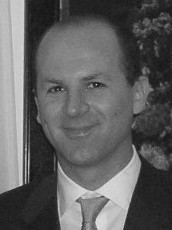}}]{Maurizio~Lualdi}
was born in Busto Arsizio, Italy in 1973. He completed his studies in Enviromental Engineering at the Politecnico di Milano, Italy. He is Associate Professor of Applied Geophysics at the Dipartimento di Ingegneria Civile e Ambientale of Politecnico di Milano. In the last years he has focused his research interests  on the development of full polarimetric 3D Georadar surveys.
\end{IEEEbiography}

\begin{IEEEbiography}[{\includegraphics[width=1in,height=1.25in,clip,keepaspectratio]{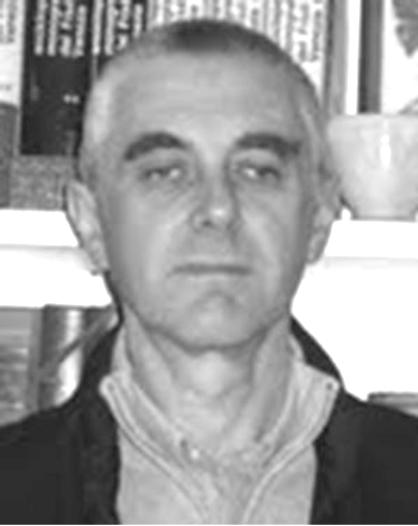}}]{Stefano~Tubaro}
(SM'01) was born in Novara, Italy, in 1957.
He completed his studies in Electronic Engineering at the Politecnico di Milano, Milan, Italy, in 1982.
He then joined the Dipartimento di Elettronica, Informazione e Bioingegneria of the Politecnico di Milano, first as a Researcher of the National Research Council, and then (in November 1991) as an Associate Professor.
Since December 2004, he has been appointed as a Full Professor of telecommunication at the Politecnico di Milano.
His current research interests include advanced algorithms for video and sound processing.
He is the author of more than 180 scientific publications on international journals and congresses and the coauthor of more than 15 patents.
In the past few years, he has focused his interest on the development of innovative techniques for image and video tampering detection and, in general, for the blind recovery of the ``processing history'' of multimedia objects.
He coordinates the research activities of the Image and Sound Processing Group at the Dipartimento di Elettronica, Informazione e Bioingegneria, Politecnico di Milano.
He had the role of Project Coordinator of the European Project ORIGAMI (A new paradigm for high-quality mixing of real and virtual) and of the research project ICT-FET-OPEN REWIND (REVerse engineering of audio-VIsual coNtent Data).
This last project was aimed at synergistically combining principles of signal processing, machine learning, and information theory to answer relevant questions on the past history of such objects.
He is a member the IEEE Multimedia Signal Processing Technical Committee and of the IEEE SPS Image Video and Multidimensional Signal Technical Committee.
He was in the organization committee of a number of international conferences including the IEEE MMSP 2004/2013, IEEE ICIP 2005, IEEE AVSS 2005/2009, IEEE ICDSC 2009, IEEE MMSP 2013, IEEE ICME 2015.
From May 2012 to April 2015, he was an Associate Editor of the IEEE TRANSACTIONS ON IMAGE PROCESSING, and is currently an Associate Editor of the IEEE TRANSACTIONS ON INFORMATION FORENSICS AND SECURITY.
\end{IEEEbiography}

\end{document}